# Computer-aided diagnosis in histopathological images of the endometrium using a convolutional neural network and attention mechanisms

Hao Sun[1,+], Xianxu Zeng[2,+], Tao Xu[3], Gang Peng[4] & Yutao Ma[1,*]

1. School of Computer Science, Wuhan University, Wuhan 430072, China.
2. Department of Pathology, the Third Affiliated Hospital of Zhengzhou University, Zhengzhou, 450052, China.
3. Zhengzhou Ultralucia Medical Technology Company Limited, Zhengzhou 450016, China.
4. Cancer Center, Union Hospital, Tongji Medical College, Huazhong University of Science and Technology, Wuhan 430022, China.

+ These authors contributed equally to this work.
* Corresponding author: Yutao Ma, School of Computer Science, Wuhan University, Wuhan 430072, China, Tel: +86 (27)6877-6081, Fax: +86 (27)6877-4590, E-mail: ytma@whu.edu.cn.

## Abstract

**Background**: Uterine cancer, also known as endometrial cancer, can seriously affect the female reproductive organs, and histopathological image analysis is the gold standard for diagnosing endometrial cancer. Computer-aided diagnosis (CADx) approaches based on traditional machine learning algorithms have been proposed to assist pathologists in interpreting histopathological images efficiently. However, due to the limited capability of modeling the complicated relationships between histopathological images and their interpretations, these CADx approaches often failed to achieve satisfying results.

**Methods**: In this study, we developed a CADx approach using a convolutional neural network (CNN) and attention mechanisms, called HIENet. Because HIENet used the attention mechanisms and feature map visualization techniques, it can provide pathologists better interpretability of diagnoses by highlighting the histopathological correlations of local (pixel-level) image features to morphological characteristics of endometrial tissue. We then evaluated the classification performance of HIENet in ten-fold cross-validation on ~3,300 hematoxylin and eosin (H&E) images (collected from ~500 endometrial specimens from October 2017 to August 2018) and external validation on additional 200 H&E images (collected from 50 randomly-selected female patients during the first quarter of 2019).

**Results**: In the ten-fold cross-validation process, the CADx approach achieved a 76.91 ± 1.17% (mean ± s. d.) classification accuracy for four classes of endometrial tissue, namely normal endometrium, endometrial polyp, endometrial hyperplasia, and endometrial adenocarcinoma. Also, HIENet achieved an area-under-the-curve (AUC) of 0.9579 ± 0.0103 with an 81.04 ± 3.87% sensitivity and 94.78 ± 0.87% specificity in a binary classification task that detected endometrioid adenocarcinoma ("Malignant"). Besides, in the external validation process, the CADx approach achieved an 84.50% accuracy in the four-class classification task, and it achieved an AUC of 0.9829 with a 77.97% (95% CI, 65.27%–87.71%) sensitivity and 100% (95% CI, 97.42%–100.00%) specificity. Moreover, positive predictive value (PPV) and negative predictive value (NPV) reached 100% (95% CI, 92.29%–100.00%) and 91.56% (95% CI, 86.00%–95.43%), respectively. The classification performance of HIENet can be further improved if directly trained as a binary classification model (also known as a binary classifier).

**Conclusion**: The proposed CADx approach, HIENet, outperformed three human experts and four end-to-end CNN-based classifiers on this small-scale dataset regarding overall classification performance. It was also able to identify some typical morphological characteristics in H&E images to provide histopathological interpretations for pathologists.

## Keywords

# Introduction

Cancer of the uterine corpus (or corpus uteri), also called uterine cancer (as opposed to cervical cancer), was the sixth most frequently diagnosed cancer among women worldwide, with an estimated 319,600 diagnosed new cases in 2012 [1]. Endometrial cancer, which arises from the endometrium of the uterus, is the most common form of uterine cancer. Therefore, endometrial cancer is sometimes known as uterine cancer in a general sense. Incidence and mortality rates for endometrial cancer in females are higher in developed countries, and the incidence rate is increasing [2]. According to a report released by the American Cancer Society in 2018 [3], endometrial cancer is the fourth most common cancer overall in the United States, accounting for 7% of all new cancer diagnoses in women. In general, if endometrial lesions can be detected early using commonly-used clinical screening and detection techniques (for example, transvaginal ultrasound [4], hysteroscopy [5], and hysterosalpingography [6]), the treatment outcome will be favorable. Moreover, the five-year survival rate for endometrial cancer after undergoing appropriate treatment is over 80% [7].

Since computer-aided diagnosis (CADx), also known as computer-aided detection (CADe), can assist doctors in efficiently analyzing and evaluating a vast number of medical images like ultrasound images and X-rays, it has been used in the radiologic diagnosis of some common cancers, including breast cancer [8], lung cancer [9], and colon cancer [10], in clinical environments. In recent years, a few researchers also developed CADx systems that were able to process hysteroscopic [11,12], ultrasound [13,14], magnetic resonance [15], and histological images [16] for early detection of endometrial cancer. However, due to the small size of training samples and limited capability of feature extraction (that is to say, heavy dependence on handcrafted image features), these CADx systems failed to achieve an adequate level of overall performance. For example, Neofytou *et al.* [12] developed a support vector machine (SVM) classification model (or called an SVM classifier) with statistical and gray-level difference statistics (GLDS) features, but only obtained an 81% accuracy on a dataset of 516 regions of interest.

Deep learning [17], one of the most exciting recent advances in machine learning, has made considerable success in many real-world applications like image recognition [18], playing the game of Go [19], and machine translation [20]. Besides, it has achieved a breakthrough in the digital image-based identification of some specific cancers or rare diseases, such as diabetic retinopathy [21,24], congenital cataract [22], skin cancer [23], and bacterial and viral pneumonia [24], showing human expert-levels of performance in the classification of diseases. Thus, the combination of CADx and deep learning has great potential to improve further the efficacy of traditional CADx systems for endometrial cancer using big data in clinical imaging.

In addition to the potential applications in radiology, other critical future applications include research fields such as pathology, especially in digital pathology based on whole-slide imaging [25] and artificial intelligence (AI). As we know, endometrial biopsy [26] with histopathological confirmation is the gold standard for diagnosing endometrial cancer [27]. Also, understanding the histopathology of endometrial cancer at the cellular level is currently recognized as the most reliable diagnosis of the disease. The demand for such applications in pathology is exploding [28] because the dearth of pathologies has become a real bottleneck to efficient, accurate, and convenient medical care in most of the developing countries in the world. For example, according to the Chinese Pathologist Association, China, as a country with a population of 1.4 billion, has only 20,000 licensed pathologists – about the same number as the United States. As a result, most pathologists in these countries are overworked and forced to examine histological sections of stained endometrial specimens as quickly as possible, which sometimes leads to reports of conflicting results or even an incorrect result. The status quo would be changed by digital pathology with the advent of deep learning that will be used to classify digitized pathology slides automatically – similar to the recent successful works in radiology for diabetic retinopathy, lung cancer, breast cancer and so on [29].

The purpose of this study is to develop a CADx approach based on deep learning to assist pathologists in efficiently evaluating histological images from endometrial tissue samples stained with hematoxylin and eosin (H&E). In addition to accurate image classification, we attempt to provide diagnostic interpretability (that is, histopathological correlation of different types of endometrial tissues with H&E image features extracted by our CADx approach) for pathologists to interpret and analyze H&E images effectively. Therefore, we expect our work may help improve the efficiency and productivity of pathologists in diagnosing endometrial diseases.

## Materials and Methods

### Sample collection
The Institutional Review Board of the Third Affiliated Hospital of Zhengzhou University approved this study. Selected female patients who were treated at the Third Affiliated Hospital of Zhengzhou University from October 2017 to August 2018 participated in the study with informed consents. Fresh endometrial specimens were collected via hysteroscopic surgery or hysterectomy, and none of the patients underwent hormone therapy, radiotherapy, or chemotherapy before surgery. Eventually, we obtained 498 endometrial tissue samples from these patients without collecting their personal information.

### Sample preparation
The entire procedure follows a standard protocol of histological reporting of endometrial cancer released by the Royal College of Pathologists (https://www.rcpath.org). Fixation is the first step to prepare a standard sample of endometrial tissue for light microscopy. Fresh endometrial specimens were promptly fixed in 10% neutral buffered formalin at room temperature for 18–24 hours. The second step is dehydration, which removes water from formalin-fixed specimens by immersing them in a series of alcohol solutions of increasing alcoholic concentration [30]. After clearing the dehydrated specimens with organic solvents, the third step in sample preparation is embedding them in melted paraffin. Once cooled, each paraffin-embedded block (1.5 x 1.5 x 0.3 cm$^3$) was sectioned into serial four $\mu m$ slices, each of which was then placed onto a slide with albumin. The fourth step of staining begins after the dissolution of the paraffin. Histological slides of endometrial tissue were stained with H&E, which is the most frequently used stain in medical diagnosis [31]. Eventually, three skilled pathologists with more than ten years of experience in pathology examined the histological slides under light microscopy and selected representative H&E slides with diagnostic results which were made unanimously.

**Digital imaging**
By using a Mixotic scanner, these selected H&E slides were scanned into high-resolution digital images that were captured at 10x or 20x magnification. To facilitate medical image analysis using deep learning algorithms, we saved the digital images in a file format of the joint photographic experts group (JPEG).

**Histopathological image processing**
For each digital image obtained, we further extracted histopathological images (640 x 480 pixels) of a lesion or a healthy tissue from the original whole-slide image using Olympus ImageView. Besides, each digital histopathological image was marked with a class label derived from the corresponding diagnosis (see the taxonomy of digital histopathological image labels).

**Taxonomy of digital histopathological image labels**

**Table 1.** A brief introduction to the experimental dataset.

| General class | Fine-grained class | Subtype | #Images |
|---|---|---|---|
| Benign | Normal Endometrium (NE) | Luteal phase | 600 |
| | | Menstrual phase | 21 |
| | | Follicular phase | 712 |
| | Endometrial Polyp (EP) | - | 636 |
| | Endometrial Hyperplasia (EH) | Simple | 516 |
| | | Complex | 282 |
| Malignant | Endometrial Adenocarcinoma (EA) | - | 535 (16.21%) |
| Total | | | 3,302 |

Postoperative pathology reports demonstrated that there were four fine-grained classes of endometrial tissue, namely the (normal) endometrium within a regular menstrual cycle, endometrial polyp, endometrial hyperplasia, and endometrial adenocarcinoma. As described in **Table 1**, normal endometrium (NE) has three subtypes defined by the phases of the menstrual cycle, namely the luteal phase, menstrual phase, and follicular phase. The three subtypes of NE have 600, 21, and

712 images, respectively. Similarly, endometrial hyperplasia (EH) has two subtypes, namely "Simple" (short for simple hyperplasia without atypia) and "Complex" (short for complex hyperplasia without atypia), including 516 and 282 images, respectively. Endometrial polyp (EP) and endometrial adenocarcinoma (EA) contain 636 and 535 images, respectively. Besides, we defined two general classes to distinguish EA ("Malignant") from NE, EP, and EH ("Benign"). In other words, such a binary classification for a patient specifies whether the patient needs surgical therapy. This dataset that consists of 3,302 pathologically proven JPEG files of digital histopathological images of the endometrium is available for download at https://doi.org/10.6084/m9.figshare.7306361.v2.

**End-to-end convolutional neural networks for comparison**

**Table 2.** Hyper-parameter settings for different CNN-based classifiers.

| Hyper-parameter | AlexNet | VGG-16 | InceptionV3 | ResNet-50 |
|---|---|---|---|---|
| Initial learning rate | 0.005 | | | |
| Learning rate delay | 0.5 ($p$=3) | | | |
| Loss function | Categorical Cross-Entropy | | | |
| Optimizer | Adam [38] was used with standard parameter values, namely $\beta_1 = 0.9$ and $\beta_2 = 0.999$. | | | |
| Batch-size | 32 | | | |

Here, $p$ (short for "patience") represents the number of epochs in a training process. If no improvement in classification accuracy on training data is achieved for $p$ epochs, the learning rate will be reduced.

As a widely-used class of deep learning algorithms, convolutional neural networks (CNNs) have proven to be successful in the field of biomedical imaging. To compare the performance of various deep learning algorithms on the experimental dataset, we trained four different classifiers using four commonly-used CNNs, namely AlexNet [32], VGG-16 [33], InceptionV3 [34], and ResNet-50 [35]. Note that these classifiers were trained end to end with different hyper-parameters. As with some previous studies [36], we took advantage of pre-trained weights on a subset of the ImageNet dataset [37] to fine-tune the four classifiers. **Table 2** presents the basic configuration settings for them.

**HIENet**

In this study, we proposed a skillfully-designed network structure, entitled *HIENet*, to perform classification tasks for histopathological images of the endometrium. **Fig. 1a** presents the overall architecture of HIENet. HIENet is designed based on a backbone network of VGG-16 and introduces two essential blocks built with the visual attention mechanism [39]. HIENet shares the same configuration setting as VGG-16. For each input image, the backbone of HIENet extracts its image features and outputs a feature map represented by a three-dimensional (3-D) matrix. The two blocks of HIENet, namely *Position Attention* and *Channel Attention*, take the feature map as an input and then produce two new feature maps of equal size. After concatenating the new feature maps and the original one into a bigger feature map, global average pooling (GAP) and flattening operations are applied to it separately to generate two new feature vectors. Then, three consecutive fully-connected layers process the concatenation of the the two feature vectors. At last, the *softmax* function of HIENet calculates a probability distribution over four predicted output classes.

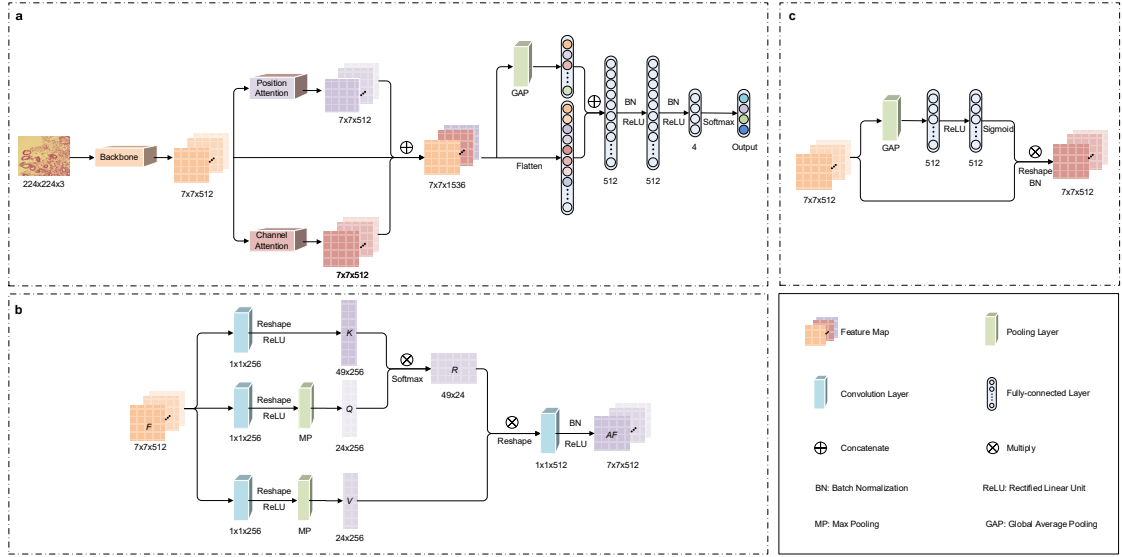

**Figure 1. The framework of HIENet.** (**a**) The overall architecture of HIENet, (**b**) Components of the Position Attention Block, (**c**) Components of the Channel Attention Block. Legends used in this figure are placed in the lower right corner. Details of HIENet refer to **Appendix S1**.

The Position Attention block (see **Fig. 1b**) introduces a self-attention mechanism, that is, non-local operation [40], to capture the relationships (more specifically, context relations) between different local areas in the input image. Recent studies [41]-[43] have demonstrated that such a type of information can improve the effects of semantic segmentation using CNNs. As mentioned above, this block takes the original feature map extracted by the backbone as an input. Three convolutional layers are trained to learn different context relations in the original feature map in parallel. Reshaping, activation, max pooling (MP), and matrix multiplication operations are then employed to deal with the intermediate results of the three convolutional layers. A detailed calculation process of the non-local operation refers to **Appendix S2**. Finally, the last convolutional layer outputs a new feature map that contains the context relations between local features extracted by the backbone.

The primary goal of the Channel Attention block (see **Fig. 1c**) is to learn the channel-wise attention, more specifically, the weights of channels. Previous studies [44],[45] have proved that different channels of feature maps extracted by CNNs imply semantic information of varying degrees of importance. Therefore, it is helpful to achieve better image classification results by leveraging such a type of information. First of all, the channel weights of the original feature map are initialized by a GAP operation. Then, two consecutive fully-connected layers are trained to learn new weights for those original features. At last, a new feature map is generated by combining the new weights and the original features.

**Image processing for model training**

To leverage the global information of each histopathological image, we used the whole image as an input to these CNN-based classifiers. At first, each input image was resized to 224 x 224 pixels (299 x 299 pixels for InceptionV3) to meet the size requirement of CNNs. Then, we utilized a commonly-used data normalization method, the standard score (also called z-score), to normalize the input images. During the training process, a random data augmentation (that is, the probability of horizontal or vertical flip is 50%) process was applied to each of the training images to provide rich information for the CNN-based classifiers. Besides, we used the batch normalization (BN) method [46] to normalize layer inputs.

**Evaluation methods**

Our evaluation of the five CNN-based classifiers' classification results used the ten-fold cross-validation method [47]. The experimental dataset was randomly partitioned into ten equal-sized subsamples, one of which was used for testing and the remaining nine subsamples for training each time. The cross-validation process was repeated ten times, with each of the ten subsamples used only once as test data. We then averaged the ten results as an overall evaluation.

To compare the difference between human and machine in image classification, three experienced pathologists (Investigators 1, 2, and 3) who were associate chief physicians from the Department of Pathology in the Third Affiliated Hospital of Zhengzhou University participated in this study. Our evaluation of the three experts' classification results used a two-step method including pre-testing and blinded-testing. Each investigator was required to make a diagnosis for each of the 50 images in the same pre-testing dataset. After the diagnosis was made, we provided instant feedback with ground truth to each investigator to help the investigator identify these digital histopathological images. The pre-testing process was repeated until each investigator was ready for the final test. Finally, the three investigators performed a blinded test and evaluated 100 label-free images (that is, a subset of one subsample of the experimental dataset in the ten-fold cross-validation process) separately and independently. For both human and machine, the likelihood that an image belongs to a general class (that is, "Benign" or "Malignant") is inferred by summing up the probabilities over all the subclasses of the general class.

**Evaluation metrics**

We quantified evaluation results using three frequently-used metrics, namely accuracy, sensitivity, and specificity. Suppose *TP*, *TN*, *FP*, and *FN* represent true positives (classified as positive correctly), true negatives (classified as negative correctly), false positives (classified as positive incorrectly), and false negatives (classified as negative incorrectly), respectively, the three metrics are defined as follows.

$$Accuracy = \frac{TP+TN}{TP+FP+TN+FN} \qquad (1)$$

$$Sensitivity = \frac{TP}{TP+FN} \qquad (2)$$

$$Specificity = \frac{TN}{TN+FP} \qquad (3)$$

Besides, we visualized the performance of the proposed HIENet and human experts using confusion matrices (also known as error matrices) [48] and receiver operating characteristic (ROC) curves [49]. In a confusion matrix, each row (or column) represents predicted results while each column (or row) represents actual classes, thus making it easy to see misclassification patterns made by them. A ROC curve is generated by plotting sensitivity against (1 – specificity) at various threshold settings. Also, the area under the ROC curve (AUC) was calculated to evaluate the performance of the five CNN-based classifiers in a binary classification task. Higher AUC values indicate better binary classification results.

**Image feature visualization**
In this study, two commonly-used visualization methods for feature maps were utilized to help pathologists understand and reason about the output of HIENet. We provided better interpretability of our approach via the correlation of pixel-level image features (that is, the concatenation of the three feature maps in **Fig. 1a**) to histopathological interpretations. Springenberg *et al.* [50] proposed a method of guided backpropagation (GB), a new variant of the "deconvolution approach," which has been used to visualize image features learned by CNNs. We also used a class activation map (CAM) [51] that generated a heat map to highlight which regions in an image were relevant to the given class.

**External Validation**
A recent report by Kim *et al.* [52] found that only 31 out of 516 published studies on the diagnostic analysis of medical images using AI algorithms performed external validation. In addition to the internal validation based on ten-fold cross-validation, we collected a new dataset of histopathological images to test the generalizability and robustness of HIENet. This dataset used for external validation has a total number of 200 images, including 74 NE images, 12 EP images, 55 EH images, and 59 EA images (29.5%). They were collected from 50 randomly-selected Chinese female patients who had been examined at the Third Affiliated Hospital of Zhengzhou University during the first quarter of 2019. The dataset is available for research at https://github.com/ssea-lab/DL4ETI.

**Code availability**
Our experiments of image classification were performed on a Dell Precision workstation T5810 with one central processing unit (CPU, Intel Xeon E5-1620 v3, 3.5 GHz) and one graphics processing unit (GPU, NVIDIA GeForce GTX 1080, 8GB).

The operating system of the workstation is Ubuntu 16.04 (64-bit). The code used to process and classify histopathological images was written in Python 3.6.5. In the meantime, we used Keras 2.1.3 (https://keras.io) and TensorFlow 1.2.0 (https://www.tensorflow.org) as the deep learning framework for the experiments. All the source code of this study is available at https://github.com/ssea-lab/DL4ETI.

# Results

## Comparison among different CNN-based classifiers

**Table 3.** CNN-based classifiers' classification results (mean ± s. d.).

| Classifier | Accuracy (%) | | Sensitivity (%) | Specificity (%) | AUC |
|---|---|---|---|---|---|
| | Four classes | B/M | B/M | B/M | B/M |
| AlexNet | $56.26 \pm 2.83$ | $83.94 \pm 3.22$ | $27.95 \pm 27.81$ | $94.41 \pm 2.71$ | $0.8735 \pm 0.0226$ |
| VGG-16 | $74.79 \pm 3.45$ | $91.28 \pm 1.23$ | $75.52 \pm 8.44$ | $94.26 \pm 0.97$ | $0.9525 \pm 0.0127$ |
| InceptionV3 | $58.24 \pm 3.54$ | $85.62 \pm 2.97$ | $36.42 \pm 29.76$ | $93.17 \pm 2.89$ | $0.8893 \pm 0.0293$ |
| ResNet-50 | $59.54 \pm 1.07$ | $86.64 \pm 1.37$ | $55.91 \pm 10.82$ | $91.25 \pm 1.72$ | $0.8819 \pm 0.0197$ |
| HIENet | $76.91 \pm 1.17$ | $93.53 \pm 0.81$ | $81.04 \pm 3.87$ | $94.78 \pm 0.87$ | $0.9579 \pm 0.0103$ |
| HIENet$_{BC}$ | — | $95.94 \pm 0.74$ | $84.45 \pm 4.79$ | $98.15 \pm 0.88$ | $0.9808 \pm 0.0094$ |

B/M: the binary classification task that distinguishes "Malignant" from "Benign." HIENet$_{BC}$ denotes that HIENet was directly trained to be a binary classifier for B/M.

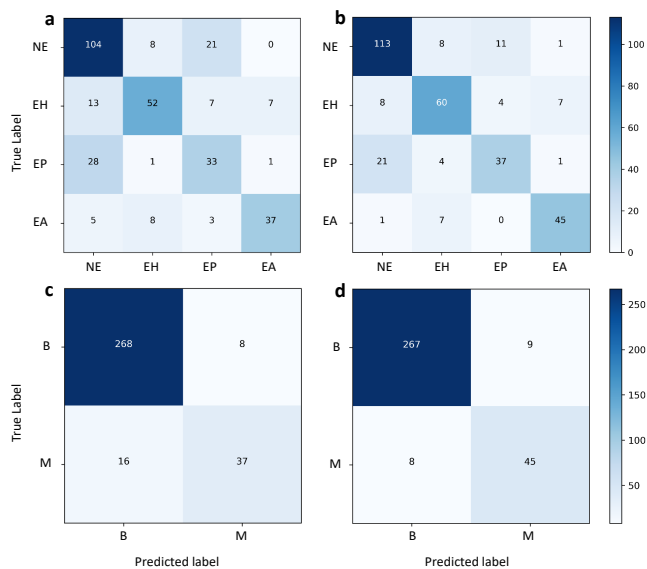

**Figure 2.** **Confusion matrices of HIENet and VGG-16 in the two classification tasks.** (**a**) VGG-16 for the four classes, (**b**) HIENet for the four classes, (**c**) VGG-16 for the two classes, (**d**) HIENet for the two classes. Cell ($i, j$) in a confusion matrix represents the probability of predicting class $j$ given that the actual label is $i$. The darker the color of the diagonal cells, the higher the correct classification rate.

We designed a four-class classification task and a binary classification task for the five CNN-based classifiers to test their performance on the experimental dataset. **Table 3** shows their classification results for the two classification tasks in the ten-fold cross-validation process. In the four-class classification task, our CADx approach, HIENet, performed the best and achieved a 76.91 ± 1.17% (mean ± s. d.) accuracy, followed by VGG-16 with a 74.79 ± 3.45% accuracy. A more detailed view of HIENet's classification accuracy values in the whole cross-validation process is shown in **Appendix S3**. The other three CNN-based classifiers performed significantly worse than HIENet and VGG-16. **Fig. 2** displays confusion matrices of HIENet and VGG-16 for one subsample of the experimental dataset in the two classification tasks. Compared with VGG-16, HIENet had lower misclassification rates in identifying NE, EH, EP, and EA, and the misclassification rates for the four general classes were decreased by 6.77%, 10.13%, 6.35%, and 15.09%, respectively.

As shown in **Table 3**, HIENet also performed the best in the binary classification task. HIENet achieved a 93.53 ± 0.81% accuracy and was slightly better than VGG-16. Generally speaking, the remaining three CNN-based classifiers worked well in the binary classification task because their AUC values were close to 0.9. It is worth noting that HIENet achieved an AUC value of 0.9579 ± 0.0103 with an 81.04 ± 3.87% sensitivity and 94.78 ± 0.87% specificity. In particular, the performance of HIENet was considerably better than those of the other four CNN-based classifiers regarding sensitivity. Moreover, **Fig. 2** shows that the misclassification rate of HIENet for "Malignant" (including EA) was decreased by 15.09% compared with VGG-16.

**Comparison between HIENet and human experts**

**Table 4. Classification results (95% CI) of human experts and HIENet.**

| | Accuracy (%) | | Sensitivity (%) | Specificity (%) | PPV (%) | NPV (%) |
|---|---|---|---|---|---|---|
| | **Four classes** | **B/M** | **B/M** | **B/M** | **B/M** | **B/M** |
| Investigator 1 | 71.00 | 89.00 | 94.74 | 87.65 | 64.29 | 98.51 |
| | (61.07–79.64) | (81.17–94.38) | (73.97–99.87) | (78.47–93.92) | (44.07–81.36) | (92.50–99.96) |
| Investigator 2 | 59.00 | 92.00 | 78.95 | 95.06 | 78.95 | 95.06 |
| | (48.71–68.74) | (84.84–96.48) | (54.43–93.95) | (87.84–98.64) | (54.43–93.95) | (87.84–98.64) |
| Investigator 3 | 58.00 | 84.00 | 94.74 | 81.48 | 54.55 | 98.51 |
| | (47.71–67.80) | (75.32–90.57) | (73.97–99.87) | (71.30–89.25) | (36.35–71.89) | (91.96–99.96) |
| Avg. (95% CI) | 62.67 | 88.33 | 89.48 | 88.06 | 65.93 | 97.36 |
| | (52.43–72.14) | (80.37–93.89) | (66.87–98.70) | (78.95–94.21) | (43.00–81.32) | (90.49–99.67) |
| HIENet (95% CI) | 76.91 | 93.53 | 81.04 | 94.78 | 82.94 | 96.53 |
| | (71.79–81.36) | (90.31–95.94) | (67.92–90.50) | (91.46–97.09) | (70.25–91.80) | (93.61–98.36) |
| HIENet$_{BC}$ (95% CI) | — | 95.94 | 84.45 | 98.15 | 90.08 | 97.04 |
| | | (93.20–97.80) | (71.87–92.94) | (95.77–99.39) | (78.44–96.67) | (94.30–98.70) |

B/M: the binary classification task that distinguishes "Malignant" from "Benign." PPV: positive predictive value ($\frac{TP}{TP+FP}$); NPV: negative predictive value ($\frac{TN}{TN+FN}$); CI: confidence interval. Note that CIs for accuracy, sensitivity, specificity, PPV, and NPV are "exact" Clopper-Pearson confidence intervals [53] at the 95% confidence level. HIENet$_{BC}$ denotes that HIENet was directly trained to be a binary classifier for B/M.

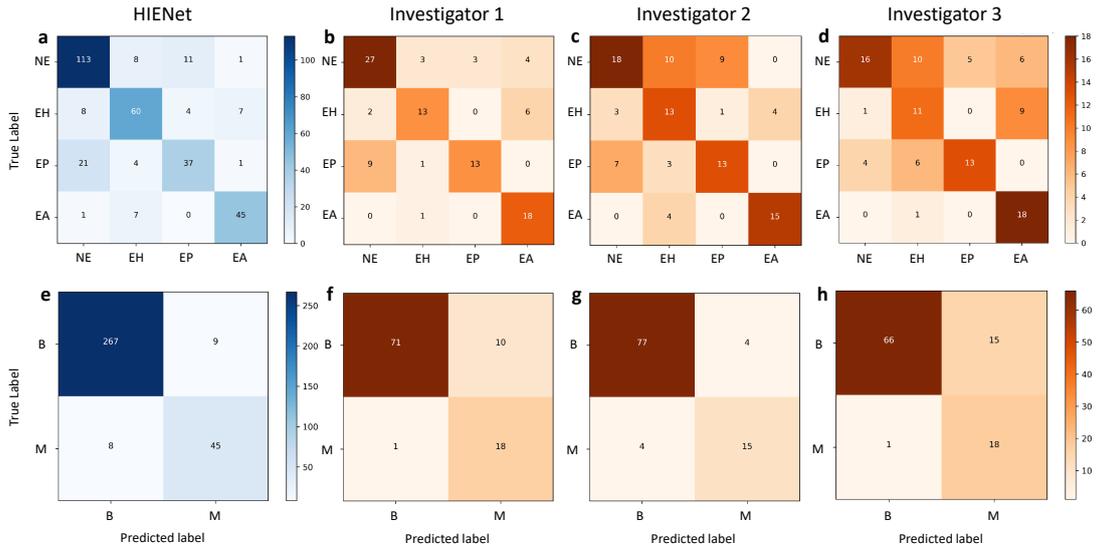

**Figure 3.** **Confusion matrices of HIENet and the three investigators in the two classification tasks.** (**a**)~(**d**): HIENet, Investigator 1, Investigator 2, and Investigator 3 for the four classes; (**e**)~(**h**): HIENet, Investigator 1, Investigator 2, and Investigator 3 for the two classes. Note that the dataset used for human blinded-testing (100 images) was a subset of the subsample of the experimental dataset used by HIENet (329 images).

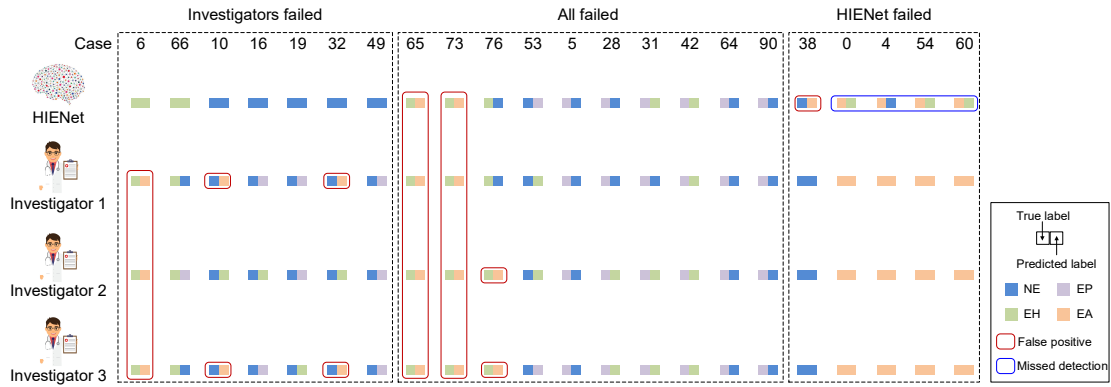

**Figure 4.** **Comparison of the performance of HIENet and individual investigators in the four-class classification task for selected cases.** The first part presents seven cases that were correctly identified by HIENet but misclassified by the three investigators, the second part shows ten cases that all subjects failed, and the third part displays five cases that were correctly classified by the three investigators but wrongly identified by HIENet.

We then compared our CADx approach (HIENet) and three human experts in the two classification tasks and presented their classification results in **Table 4**. In the four-class classification task, the three investigators achieved, on average, a 62.67% accuracy, which was ~14% lower than that of HIENet. Even for the best investigator (Investigator 1) with the highest accuracy value, the investigator's accuracy was ~6% lower than that of HIENet. **Fig. 3** displays confusion matrices of HIENet and the three investigators in the two classification tasks. Although the first investigator performed the best among the three investigators, Investigator 1's misclassification rates of NE,

EH, and EP were ~12%, ~14%, and ~2% higher than those of HIENet, respectively. Instead, this investigator missed only one out of 19 EA images. In particular, there are seven cases (that is, five NE images and two EH images) that were correctly identified by HIENet but misclassified by all the three investigators (see the first part in **Fig. 4**), and the three investigators reported seven false positives in three cases.

As shown in **Table 4**, in the binary classification task, HIENet's accuracy was ~5% higher than the average level of the three investigators. Compared with the best investigator (Investigator 2) with the highest accuracy value, HIENet's accuracy and sensitivity were ~1.5% and ~2% higher than those of the investigator, respectively. Also, HIENet's specificity was very close to the diagnostic level of the best investigator. Moreover, only nine "Benign" images (out of 276 images) were misclassified by HIENet as "Malignant," three of which refer to case numbers 38, 65, and 73 in **Fig. 4**. The misclassification rate of HIENet for "Benign" (including NE, EP, and EH) was only 3.26%, suggesting that HIENet did have a stronger ability to reject "Benign" images without a condition correctly. **Fig. 5** further depicts a ROC curve of HIENet with an AUC value of 0.9559 in the binary classification task. It is evident from **Fig. 5** that HIENet outperformed two investigators who were denoted by solid (blue) circles lying below the (black cyan) ROC curve, and that HIENet and the remaining investigator were evenly matched in this classification task. Therefore, the average level of the three investigators, which was denoted by the (red) square, dropped below the ROC curve. Although HIENet achieved an AUC value of 0.9579 ± 0.0103, its sensitivity was ~8% lower than the average sensitivity of the three investigators due to a minimal number of EA images.

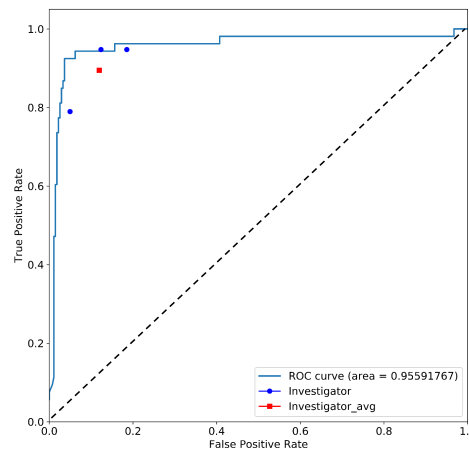

**Figure 5. ROC curve of HIENet in the binary classification task.** The X-axis represents the false positive rate (1 − specificity), while the Y-axis represents the true positive rate (sensitivity). The (black cyan) ROC curve denotes the performance of HIENet, each solid (blue) circle denotes the performance of a human expert, and the (red) square denotes the average level of the three human experts. Points above the diagonal with a dotted line indicate classification results better than random; that is, the AUC value is 0.5.

Besides these commonly-used evaluation metrics, we also provided two useful metrics, namely the positive predictive value (PPV) and negative predictive value (NPV), to evaluate the classification results. According to the last column in **Table 4**, HIENet was indeed comparable to the three investigators regarding the average NPV value. Moreover, HIENet was far better than them in terms of the average PPV value. For example, the PPV value of HIENet was ~4% higher than that of the best investigator (Investigator 2). In summary, HIENet could provide diagnoses with accuracy comparable to and in some cases better than human experts in this classification task.

## Visualization of pixel-level morphological features

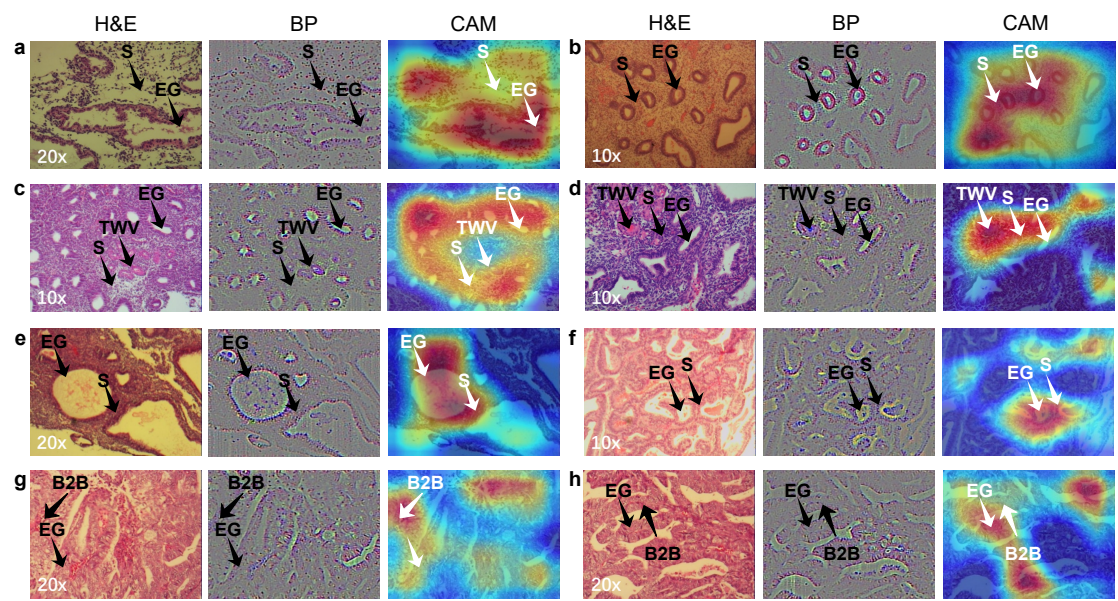

**Figure 6.** **Visualization of pixel-level morphological characteristics in H&E images extracted by our approach for four types of endometrial tissue.** (**a**)&(**b**): NE; (**c**)&(**d**): EP; (**e**)&(**f**): EH; (**g**)&(**h**): EA. The three panels in each row display H&E image, GB map, and CAM map, respectively. GB and CAM maps visualize pixel-level morphological representations learned by HIENet. EG: endometrial gland; S: stroma; TWV: thick-walled vessel; B2B: back to back. Magnification: 10x and 20x.

**Fig. 6a** presents a healthy endometrial tissue in the luteal phase. The left H&E image showed a focus of an irregular shaped endometrial gland surrounded by loose endometrial stroma; moreover, the lumen of the gland appeared to have the "sawtooth" serrated contour. Both the BP map and CAM map captured the serrated contour of this gland. The H&E image in **Fig. 6b** demonstrates an example of NE in the follicular phase. Several simple tubular glands scattered in an orderly manner, and the density of the surrounding endometrial stroma was relatively uniform. The gland-to-stroma ratio was, overall, smaller than 1:1 due to stromal proliferation. It is evident from **Fig. 6b** that the BP map and CAM map captured the natural contour of these simple tubular glands. Besides, the CAM map highlighted a small piece of tissue uniformly composed of simple tubular glands and endometrial stroma.

**Fig. 6c** and **6d** display two examples of EP. Polyps, which arise from the endometrial layer of the uterus, may cause abnormal bleeding in women during perimenopause and after menopause. Although a few endometrial glands in **Fig. 6c** looked like those simple tubular glands in **Fig. 6b**, there were thick-walled vessels (TWVs) in the fibrotic stroma, which is a morphological feature commonly found in EP [54]. In addition to the above characteristic, glandular architectural abnormality in the form of dilated glands with irregular shapes was also found in the H&E image in **Fig. 6d**. In general, the BP and CAM maps captured the collection of TWVs and the surrounding fibrotic stroma. In particular, the CAM map in **Fig. 6d** emphasized the cluster of blood vessels in a large polyp.

EH, which is an abnormal glandular proliferation of the endometrium, has become a risk factor for the development or sometimes co-existence of endometrial cancer. **Fig. 6e** demonstrates an example of simple hyperplasia without atypia. The H&E image in **Fig. 6e** shows a focus of a dilated gland with marked cystic expansion, though its appearance looked normal. Besides, the gland-to-stroma ratio increased and was higher than the observed value in normal endometrium (see **Fig. 6b**). Compared with the BP map, the CAM map had a more exceptional ability to highlight the above changes in the glandular lumen as well as in the endometrial stroma between the dilated gland and its lower-right neighbor. Besides, the H&E image in **Fig. 6f** displays an example of complex hyperplasia without atypia, showing crowding of glands that appeared disorganized and had luminal outpouching. It is evident from **Fig. 6f** that the CAM map emphasized such a structural alteration of glands, which is one of the primary morphological characteristics that distinguish complex from simple hyperplasia [55].

**Fig. 6g** and **6h** present two examples of invasive endometrial cancer (more specifically, adenocarcinoma). Clear contours of individual glands appeared to be almost lost. Although the two well-differentiated endometrial tumor samples appeared on a background of complex hyperplasia, they were identified by the finding of "back-to-back" (B2B) glands with little intervening stroma. The CAM maps in **Fig. 6g** and **6h** highlighted such a marked morphological feature of adenocarcinoma [56].

**External validation result**

In the four-class classification and binary classification tasks, our CADx approach, HIENet, achieved 84.50% and 93.50% accuracy, respectively, on the dataset used for external validation. In particular, it achieved an AUC of 0.9829 (see **Fig. 7c**) with a 77.97% (95% CI, 65.27%–87.71%) sensitivity and 100% (95% CI, 97.42%–100.00%) specificity. Moreover, PPV and NPV values reached 100% (95% CI, 92.29%–100.00%) and 91.56% (95% CI, 86.00%–95.43%), respectively. This classification result matches well with the overall performance of HIENet in the ten-fold cross-validation process. As shown in **Fig. 7b**, the misclassification rate of HIENet for

"Benign" was zero, indicating that it was indeed able to identify H&E images of benign endometrial tissues correctly.

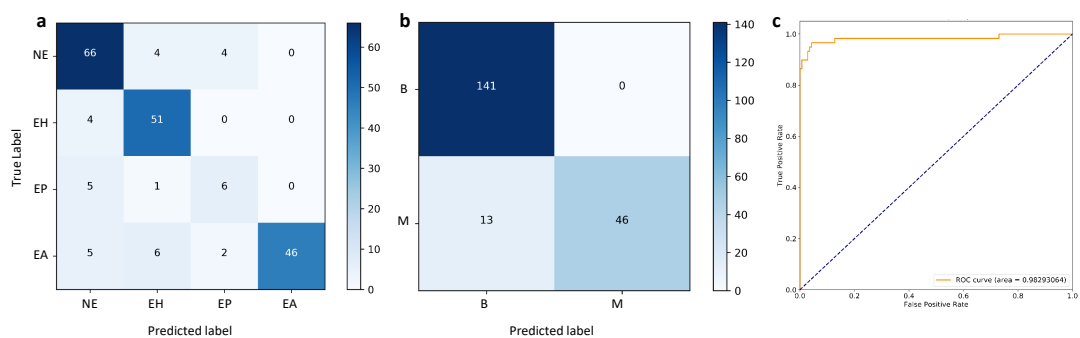

**Figure 7. External validation result of HIENet on a new dataset composed of 200 images.** (**a**): Confusion matrix of four-class classification (**b**): Confusion matrix of binary classification; (**c**): ROC curve of the binary classification task.

## Discussion

As we know, training a high-quality machine learning classifier for a specific cancer classification task via deep learning always requires a massive amount of labeled image data. Due to the high cost of the image annotation work in both time and effort, as well as the protection of patient privacy, there exist few datasets of endometrial images available to the public. For example, there are only two datasets of radiologic images for endometrial carcinoma on the website of the Cancer Imaging Archive (https://www.cancerimagingarchive.net), a public open-access database of medical images for cancer research. The shortage of labeled medical images is thus the primary challenge in endometrial image analysis using deep learning [29]. In this study, we spent one year collecting and annotating 3,500 H&E images of endometrial specimens. Although this dataset was small, our CADx approach (HIENet) performed well in both the ten-fold cross-validation for 3,300 images and the external validation for 200 images. We plan to collect more tissue samples of the endometrium in the future to enrich this dataset. The performance of HIENet (especially sensitivity) will, predictably, improve along with the increasing volume of labeled image data (especially of endometrial cancer images).

The biggest concern of this study is that HIENet had a relatively higher false negative rate than the three human experts. In particular, a few EA images had been misclassified by HIENet into EH (see **Fig. 3a** and **Fig. 7a**). The main reasons are two-fold. First, as mentioned above, the number of EA and "Complex" EH images is insufficient to train HIENet to distinguish the difference between the two types of images precisely. Second, because all the H&E images in this dataset obtained under low magnification, HIENet had to identify different types of images without segmentation according to morphological characteristics. However, a few subtle features of EA need to be detected at the cellular level [54,56]. Under these circumstances, the three human experts had more experience in utilizing only morphological characteristics to diagnose EA cases than our approach. As shown in **Fig. 8**, our approach was confused by three EA images (see case numbers 0, 54, and 60 in **Fig. 4**) that have many irregular glands scattered in a disorderly manner. Also, the regions highlighted by HIENet were very different from those recognized by the three investigators. This result implies that our approach trained by a small-scale image dataset, in some cases, failed to capture certain subtle features that can distinguish similar morphological characteristics between EA and EH.

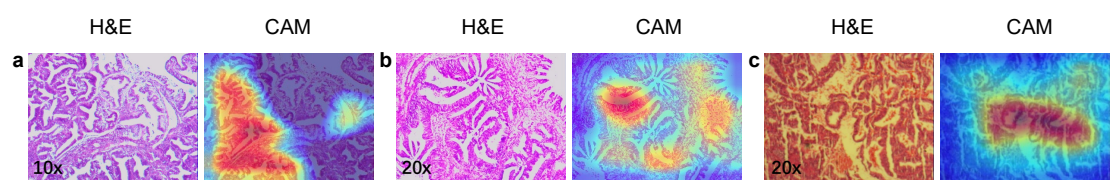

**Figure 8. Complicated cases misclassified by HIENet.** (**a**): Case No. 0; (**b**): Case No. 54; (**c**): Case No. 60. Our CADx approach misclassified the three EA cases (see the right part in **Fig. 4**) into EH, but all the three investigators diagnosed them correctly. The two panels in each row display H&E image and CAM map, respectively. The CAM maps visualize pixel-level morphological representations learned by HIENet. Magnification: 10x and 20x.

Although the proposed CADx approach cannot be applied directly to the clinical environment, it has vast potential to be used in a man-machine collaboration pattern for grading diagnosis in endometrial diseases. This pattern will make full use of their respective advantages. HIENet can be trained as a binary classifier (denoted as *HIENet*$_{BC}$) to screen out negative samples efficiently, as well as to provide suspected lesion areas in each possible positive sample image for pathologists. As shown in the last rows in **Tables 3** and **4**, in the ten-fold cross-validation process, HIENet$_{BC}$'s sensitivity and specificity were increased by ~3.4% and ~3.4%, respectively, compared with those of HIENet. The former's PPV value was ~7.1% higher than that of the latter. Besides, in the external validation, HIENet$_{BC}$ achieved an AUC of 0. 9971 with an 86.44% sensitivity (95% CI, 75.02%–93.96%) and 100% specificity (95% CI, 97.42%–100.00%). Its sensitivity was ~8.5% higher than that of HIENet. Moreover, its PPV and NPV values reached 100% (95% CI, 92.29%–100.00%) and 94.63% (95% CI, 89.69%–97.65%), respectively. Confusion matrices of HIENet$_{BC}$ corresponding to **Fig. 3e** and **Fig. 7b** are shown in **Appendix S4**. Once HIENet$_{BC}$ generates a report of binary classification results, pathologists can quickly review the report and then focus on those possible positive samples. We expect that such an AI-enabled pattern could facilitate the diagnosis process of endometrial cancers and help increase the productivity of pathologists.

Our future work to realize this pattern in the clinical environment includes three aspects. First, we will collect more image data in volume and type to enrich this dataset. For example, this dataset does not contain any image of atypical endometrial hyperplasia, which is a premalignant condition of the endometrium. The enrichment of H&E images is useful for the accurate identification of more types of endometrial diseases. Second, we will design an interactive framework of human (more specifically, expert)-in-the-loop [57,58] for HIENet. In this framework, our CADx approach collaborates with skilled pathologists and learns how they make a pathologic diagnosis on malignant endometrial lesions, which may help make HIENet both more effective and approachable. In some cases, HIENet needs to learn how to leverage human expert's feedback to correct its wrong decisions. Third, a few machine learning techniques, such as deep reinforcement learning [59] and transfer learning [60], will be applied to HIENet to enhance the self-learning ability that can convert human expert's knowledge and skill into machine-readable representation and teach itself.

## Conclusion

In summary, we developed a CADx approach to histopathological images of endometrial diseases using a convolutional neural network and attention mechanisms. The CADx approach, called HIENet, was shown to be effective for binary and multi-class classification tasks on a small-scale dataset composed of 3,500 H&E images in ten-fold cross-validation and external validation, demonstrating better classification accuracies than three associate chief physicians. By using attention mechanisms and a class activation map, HIENet can also identify and highlight morphological characteristics in H&E images to provide better interpretability (that is, histopathological correlation of pixel-level H&E image features) for pathologists. Considering the advantages mentioned above, HIENet holds the potential to be used in a human-machine collaboration pattern for grading diagnosis in endometrial diseases, which may help increase the productivity of pathologists.

## References

1. Torre, L. A. *et al*. Global cancer statistics, 2012. *CA: A Cancer Journal for Clinicians* **65**(2), 87–108 (2015). doi:10.3322/caac.21262
2. Morice, P., Leary, A., Creutzberg, C., Abu-Rustum, N. & Darai, E. Endometrial cancer. *The Lancet* **387**(10023), 1094–1108 (2016). doi:10.1016/S0140-6736(15)00130-0
3. Siegel, R. L., Miller, K. D. & Jemal, A. Cancer statistics, 2018. *CA: A Cancer Journal for Clinicians* **68**(1), 7–30 (2018). doi:10.3322/caac.21442
4. Jacobs, I. *et al*. Sensitivity of transvaginal ultrasound screening for endometrial cancer in postmenopausal women: a case-control study within the UKCTOCS cohort. *The Lancet Oncology* **12**(1), 38–48 (2011). doi:10.1016/S1470-2045(10)70268-0
5. Clark, T. J. *et al*. Accuracy of hysteroscopy in the diagnosis of endometrial cancer and hyperplasia: a systematic quantitative review. *JAMA* **288**(13), 1610–1621 (2002). doi:10.1001/jama.288.13.1610

6. Dalfó, A. R. *et al*. Diagnostic value of hysterosalpingography in the detection of intrauterine abnormalities: a comparison with hysteroscopy. *American Journal of Roentgenology* **183**(5), 1405–1409 (2004). doi:10.2214/ajr.183.5.1831405

7. Nicolaije, K. A. *et al*. Follow-up practice in endometrial cancer and the association with patient and hospital characteristics: A study from the population-based PROFILES registry. *Gynecologic Oncology* **129** (2), 324–331 (2013). doi:10.1016/j.ygyno.2013.02.018

8. Gilbert, F. J. *et al*. Single reading with computer-aided detection for screening mammography. *The New England Journal of Medicine* **359**, 1675–1684 (2008). doi:10.1056/NEJMoa0803545

9. Li, F. *et al*. Computer-aided detection of peripheral lung cancers missed at CT: ROC analyses without and with localization. *Radiology* **237**(2), 684–690 (2005). doi:10.1148/radiol.2372041555

10. Halligan, S. *et al*. Computed tomographic colonography: assessment of radiologist performance with and without computer-aided detection. *Gastroenterology* **131**(6), 1690–1699 (2006). doi:10.1053/j.gastro.2006.09.051

11. Vlachokosta, A. A. *et al*. Classification of hysteroscopical images using texture and vessel descriptors. *Medical & Biological Engineering & Computing* **51**, 859–867 (2013). doi:10.1007/s11517-013-1058-1

12. Neofytou, M. S. *et al*. Computer-aided diagnosis in hysteroscopic imaging. *IEEE Journal of Biomedical and Health Informatics* **19**(3), 1129–1136 (2015). doi:10.1109/JBHI.2014.2332760

13. Wu, J. Y. *et al*. Quantitative analysis of ultrasound images for computer-aided diagnosis. *Journal of Medical Imaging* **3**(1), 014501 (2016). doi.org:10.1117/1.JMI.3.1.014501

14. Konrad, J., Merck, D., Wu, J. Y., Tuomi, A. & Beland, M. Improving ultrasound detection of uterine adenomyosis through computational texture analysis. *Ultrasound Quarterly* **34**(1), 29–31 (2018). doi:10.1097/RUQ.0000000000000322

15. Ueno, Y. *et al*. Endometrial carcinoma: MR imaging-based texture model for preoperative risk stratification—a preliminary analysis. *Radiology* **284**(3), 748–757 (2017). doi:10.1148/radiol.2017161950

16. Vlachokosta, A. A., Asvestas, P. A., Matsopoulos, G. K., Kondi-Pafiti, A. & Vlachos, N. Classification of histological images of the endometrium using texture features. *Analytical and Quantitative Cytopathology and Histopathology* **35**(2), 105–113 (2013). PMID:23700719

17. LeCun, Y., Bengio, Y. & Hinton, G. E. Deep learning. *Nature* **521**(7553), 436–444 (2015). doi:10.1038/nature14539

18. He, K., Zhang, X., Ren, S. & Sun, J. Delving deep into rectifiers: surpassing human-level performance on ImageNet classification. *Proceedings of the 2015 IEEE International Conference on Computer Vision (ICCV'15)*, 1026–1034 (2015). doi:10.1109/ICCV.2015.123

19. Silver, D. *et al*. Mastering the game of Go with deep neural networks and tree search. *Nature* **529**(7587), 484–489 (2016). doi:10.1038/nature16961

20. Vaswani, A. *et al*. Attention is all you need. *Proceedings of the 30$^{th}$ Annual Conference on Neural Information Processing Systems (NIPS'17)*, 6000–6010 (2017). http://papers.nips.cc/paper/7181-attention-is-all-you-need

21. Gulshan, V. *et al*. Development and validation of a deep learning algorithm for detection of diabetic retinopathy in retinal fundus photographs. *JAMA* **316**(22), 2402–2410 (2016). doi:10.1001/jama.2016.17216

22. Long, E. *et al*. An artificial intelligence platform for the multihospital collaborative management of congenital cataracts. *Nature Biomedical Engineering* **1**, 0024 (2017). doi:10.1038/s41551-016-0024

23. Esteva, A. *et al*. Dermatologist-level classification of skin cancer with deep neural networks. *Nature* **542**(7639), 115–118 (2017). doi:10.1038/nature21056

24. Kermany, D. S. *et al*. Identifying Medical Diagnoses and Treatable Diseases by Image-Based Deep Learning. *Cell* **172**(5), 1122–1131.e9 (2018). doi:10.1016/j.cell.2018.02.010

25. Ghaznavi, F., Evans, A., Madabhushi, A. & Feldman, M. Digital imaging in pathology: whole-slide imaging and beyond. *Annual Review of Pathology: Mechanisms of Disease* **8**, 331–359 (2013). doi:10.1146/annurev-pathol-011811-120902

26. Noyes, R. W., Hertig, A. T. & Rock, J. Dating the endometrial biopsy. *American Journal of Obstetrics and Gynecology* **122**(2), 262–263 (1975). doi:10.1016/S0002-9378(16)33500-1

27. Revel, A. & Shushan, A. Investigation of the infertile couple: Hysteroscopy with endometrial biopsy is the gold standard investigation for abnormal uterine bleeding. *Human Reproduction* **17**(8), 1947–1949 (2002). doi:10.1093/humrep/17.8.1947

28. Jha, S. & Topol, E. J. Adapting to artificial intelligence: radiologists and pathologists as information specialists. *JAMA* **316**(22), 2353–2354 (2016). doi:10.1001/jama.2016.17438

29. Litjens, G. *et al*. A survey on deep learning in medical image analysis. *Medical Image Analysis* **42**, 60–88 (2017). doi:10.1016/j.media.2017.07.005

30. Glauert, A. M. Fixation, Dehydration and Embedding of Biological Specimens (1st Edition). Amsterdam: Elsevier Science, 1984. ISBN: 978-0720442571

31. Kiernan, J. Histological and Histochemical Methods: Theory and Practice (4th Edition). New York: Cold Spring Harbor Laboratory Press, 2008. ISBN: 978-1904842422

32. Krizhevsky, A., Sutskever, I. & Hinton, G. E. ImageNet classification with deep convolutional neural networks. *Proceedings of the 25th International Conference on Neural Information Processing Systems (NIPS'12)*, 1097–1105 (2012). https://papers.nips.cc/paper/4824-imagenet-classification-with-deep-convolutional-neural-networks

33. Simonyan, K. & Zisserman, A. Very deep convolutional networks for large-scale image recognition. *Computing Research Repository*, arXiv:1409.1556 (2014). https://arxiv.org/abs/1409.1556

34. Szegedy, C., Vanhoucke, V., Ioffe, S., Shlens, J. & Wojna, Z. Rethinking the Inception Architecture for Computer Vision. *Proceedings of the 2016 IEEE Conference on Computer Vision and Pattern Recognition (CVPR'16)*, 2818–2826 (2016). doi:10.1109/CVPR.2016.308

35. He, K., Zhang, X., Ren, S. & Sun, J. Deep Residual Learning for Image Recognition. *Proceedings of the 2016 IEEE Conference on Computer Vision and Pattern Recognition (CVPR'16)*, 770–778 (2016). doi:10.1109/CVPR.2016.90

36. Tajbakhsh N. *et al*. Convolutional neural networks for medical image analysis: full training or fine tuning? *IEEE Transactions on Medical Imaging* **35**(5), 1299–1312 (2016). doi:10.1109/TMI.2016.2535302

37. Russakovsky, O. *et al*. ImageNet Large Scale Visual Recognition Challenge. *International*

*Journal of Computer Vision* **115**(3), 211–252 (2015). doi:10.1007/s11263-015-0816-y

38. Kingma, D. P. & Ba, J. Adam: A Method for Stochastic Optimization. *Computing Research Repository*, arXiv:1412.6980 (2014). https://arxiv.org/abs/1412.6980

39. Tsotsos, J. K. *et al*. Modeling Visual Attention via Selective Tuning. *Artificial Intelligence* **78**(1-2), 507–545 (1995). doi:10.1016/0004-3702(95)00025-9

40. Wang, X., Girshick, R. B., Gupta A. & He, K. Non-Local Neural Networks. *Proceedings of the 2018 IEEE Conference on Computer Vision and Pattern Recognition (CVPR'18)*, 7794–7803 (2018). doi:10.1109/CVPR.2018.00813

41. Huang, Z. *et al*. CCNet: Criss-Cross Attention for Semantic Segmentation. *Computing Research Repository*, arXiv:1811.11721 (2018). https://arxiv.org/abs/1811.11721

42. Fu, J., Liu, J., Tian, H., Fang, Z. & Lu, H. Dual Attention Network for Scene Segmentation. *Computing Research Repository*, arXiv:1809.02983 (2018). https://arxiv.org/abs/1809.02983

43. Yuan, Y. & Wang, J. OCNet: Object Context Network for Scene Parsing. *Computing Research Repository*, arXiv:1809.00916 (2018). https://arxiv.org/abs/1809.00916

44. Hu, J. Shen, L. & Sun, G. Squeeze-and-Excitation Networks. *Proceedings of the 2018 IEEE Conference on Computer Vision and Pattern Recognition (CVPR'18)*, 7132–7141 (2018). doi:10.1109/CVPR.2018.00745

45. Chen, L. *et al*. SCA-CNN: Spatial and Channel-Wise Attention in Convolutional Networks for Image Captioning. *Proceedings of the 2017 IEEE Conference on Computer Vision and Pattern Recognition (CVPR'17)*, 6298–6306 (2017). doi:10.1109/CVPR.2017.667

46. Ioffe, S. & Szegedy, C. Batch Normalization: Accelerating Deep Network Training by Reducing Internal Covariate Shift. *Proceedings of the 32$^{nd}$ International Conference on Machine Learning (ICML'15)*, 448–456 (2015). http://proceedings.mlr.press/v37/ioffe15.html

47. Kohavi, R. A Study of Cross-Validation and Bootstrap for Accuracy Estimation and Model Selection. *Proceedings of the 4th International Joint Conference on Artificial Intelligence (IJCAI'95)*, 1137–1145 (1995). http://www.ijcai.org/Proceedings/95-2/Papers/016.pdf

48. Stehman, S. V. Selecting and interpreting measures of thematic classification accuracy. *Remote Sensing of Environment* **62**(1), 77–89 (1997). doi:10.1016/S0034-4257(97)00083-7

49. Hanley, J. A. & McNeil, B. J. The meaning and use of the area under a receiver operating characteristic (ROC) curve. *Radiology* **143**(1), 29–36 (1982). doi:10.1148/radiology.143.1.7063747

50. Springenberg, J. T. *et al*. Striving for simplicity: The all convolutional net. *Computing Research Repository*, arXiv:1412.6806 (2014). https://arxiv.org/abs/1412.6806

51. Zhou, B., Khosla, A., Lapedriza, A., Oliva, A. & Torralba, A. Learning Deep Features for Discriminative Localization. *Proceedings of the 2016 IEEE Conference on Computer Vision and Pattern Recognition (CVPR'16)*, 2921–2929 (2016). doi:10.1109/CVPR.2016.319

52. Kim, D. W. *et al*. Design Characteristics of Studies Reporting the Performance of Artificial Intelligence Algorithms for Diagnostic Analysis of Medical Images: Results from Recently Published Papers. *Korean Journal of Radiology* **20**(3), 405-410 (2019). doi:10.3348/kjr.2019.0025

53. Agresti, A. & Coull, B. A. Approximate is better than "Exact" for interval estimation of binomial proportions. *The American Statistician* **52**(2), 119–126 (1998). doi:10.2307/2685469

54. Savelli, L. *et al*. Histopathologic features and risk factors for benignity, hyperplasia, and

cancer in endometrial polyps. *American Journal of Obstetrics and Gynecology* **188**(4), 927–931 (2003). doi:10.1067/mob.2003.247

55. Horn, L.-C., Meinel, A., Handzel, R. & Einenkel, J. Histopathology of endometrial hyperplasia and endometrial carcinoma: An update. *Annals of Diagnostic Pathology* **11**(4), 297–311 (2007). doi:10.1016/j.anndiagpath.2007.05.002

56. Saso, S. *et al*. Endometrial cancer. *BMJ* **343**, d3954 (2011). doi:10.1136/bmj.d3954

57. Holzinger, A. Interactive machine learning for health informatics: when do we need the human-in-the-loop? *Brain Informatics* **3**(2): 119–131 (2016). doi:10.1007/s40708-016-0042-6

58. Zanzotto, F. M. Viewpoint: Human-in-the-loop Artificial Intelligence. *Journal of Artificial Intelligence Research* **64**, 243–252 (2019). doi:10.1613/jair.1.11345

59. Mnih, V., Kavukcuoglu, K. & Silver, D. Human-level control through deep reinforcement learning. *Nature* **518**(7540), 529–533 (2015). doi:10.1038/nature14236

60. Pan, S. & Yang, Q. A Survey on Transfer Learning. *IEEE Transactions on Knowledge and Data Engineering* **22**(10), 1345–1359 (2010). doi:10.1109/TKDE.2009.191

## Data citation

Zeng, X., Sun, H. & Ma, Y. A histopathological image dataset for endometrial disease diagnosis. figshare, https://doi.org/10.6084/m9.figshare.7306361.v2 (2018).

## Acknowledgments

We appreciate Chaoya Zhu and Jingjing Wei for their valuable work of diagnosis very much. This work was partially supported by the National Basic Research Program of China (Grant No. 2014CB340404) and the Medical Science and Technology projects of China (Grant Nos. 201503117 and 161100311100).

## Author Contributions

X.Z. collected and processed endometrial specimens, and she also made diagnoses based on histological sections of tissue samples stained with H&E. H.S. designed HIENet and implemented CNN-based classifiers, and he tested their performance on this dataset using ten-fold cross-validation and external validation. T.X. analyzed and visualized the experimental results. P.G. provided suggestions on endometrial adenocarcinoma diagnosis and statistical analysis, and he reviewed the paper. Y.M. designed the whole framework of experiments and wrote the paper.

## Additional Information

Competing interests: The authors declare no competing financial interests.

# Appendix S1

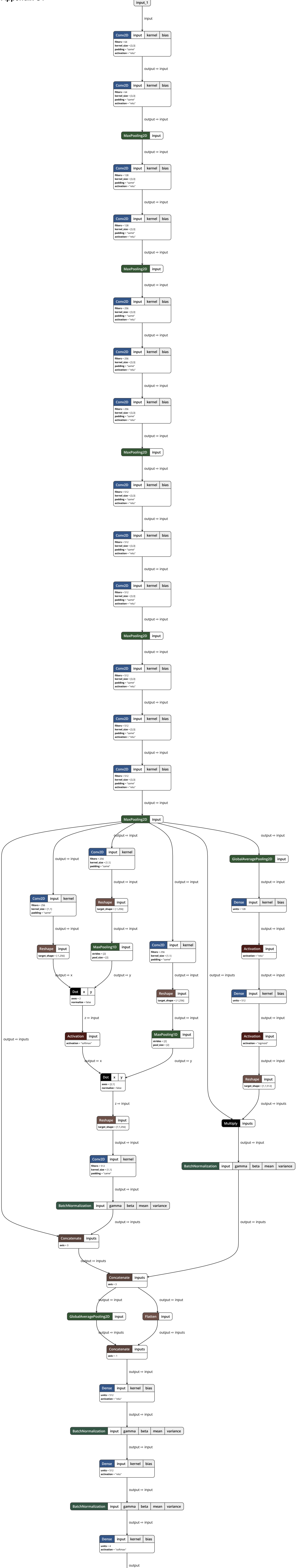

# Appendix S2

As shown in **Fig. 1b**, the Position Attention block takes the original feature map $F \in \mathbb{R}^{w \times h \times c}$ as an input. Here, parameters $w$, $h$ and $c$ denote the width, height and channel of a feature map, respectively. Each feature in a feature map contains some local information from the input image. Feature map $K \in \mathbb{R}^{wh \times c}$ is generated by the convolution operation and matrix reshaping, formally formulated as

$$K = Reshape(Conv\_K(F)), \tag{1}$$

where $Reshape$ is a function to change the shape of the input matrix while keeping its original data and $Conv\_K$ is the 1x1 convolution operation.

Feature maps $Q \in \mathbb{R}^{\lfloor \frac{wh}{2} \rfloor \times c}$ and $V \in \mathbb{R}^{\lfloor \frac{wh}{2} \rfloor \times c}$ are generated by the convolution operation, matrix reshaping, and max pooling (MP), formally formulated as

$$Q = MP(Reshape(Conv\_Q(F))), \tag{2}$$

$$V = MP(Reshape(Conv\_V(F))), \tag{3}$$

where $Conv\_Q$ and $Conv\_V$ are the 1x1 convolution operation.

Then, the context relations between features are calculated by the non-local operation in the embedded Gaussian version [40], formally formulated as

$$r_{ij} = \frac{\exp(\mathbf{k}_i \cdot \mathbf{q}_j^T)}{\sum_{\forall j} \exp(\mathbf{k}_i \cdot \mathbf{q}_j^T)}, \tag{4}$$

where $r_{ij}$ represents the correlation between the $i^{th}$ feature ($\mathbf{k}_i$) in $K$ and the $j^{th}$ feature ($\mathbf{q}_j$) in $Q$. Note that $\mathbf{k}_i \cdot \mathbf{q}_j^T$ is a dot-product similarity and the denominator of Eq. (4) is a normalization factor. By applying the $softmax$ function to this matrix $R \in \mathbb{R}^{wh \times \lfloor \frac{wh}{2} \rfloor}$ and multiplying it by $V$, we can obtain the weights of position attention between different local features.

$$Attention = Softmax(KQ^T)V. \tag{5}$$

Finally, a new feature map of the same size as $F$ ($AF \in \mathbb{R}^{w \times h \times c}$), which contains the context relation information, is generated according to Eq. (6).

$$AF = BN(Conv\_A(Reshape(Attention))), \tag{6}$$

where $Conv\_A$ is the 1x1 convolution operation and $BN$ is the Batch Normalization operation.

# Appendix S3

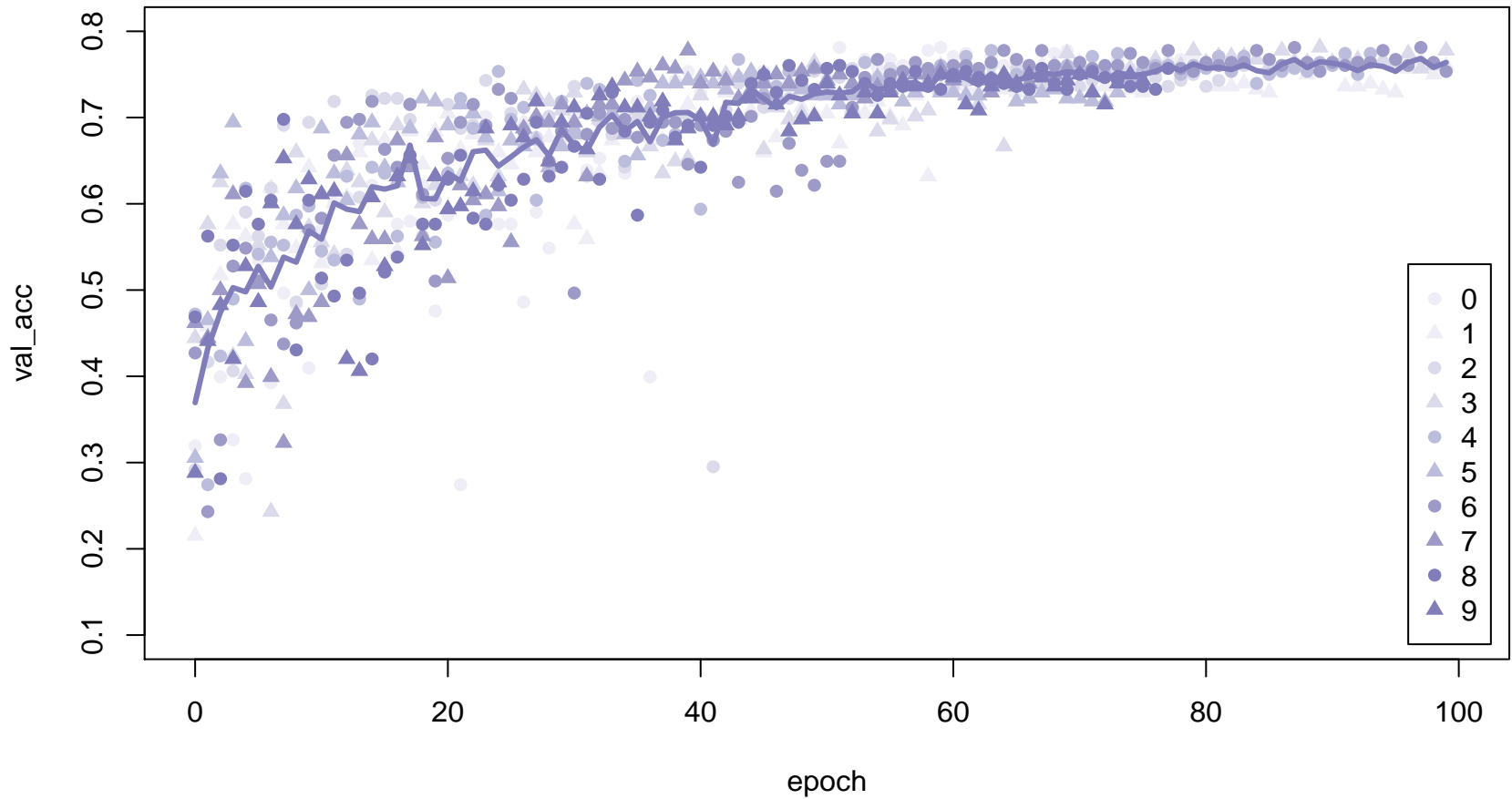

# Appendix S4

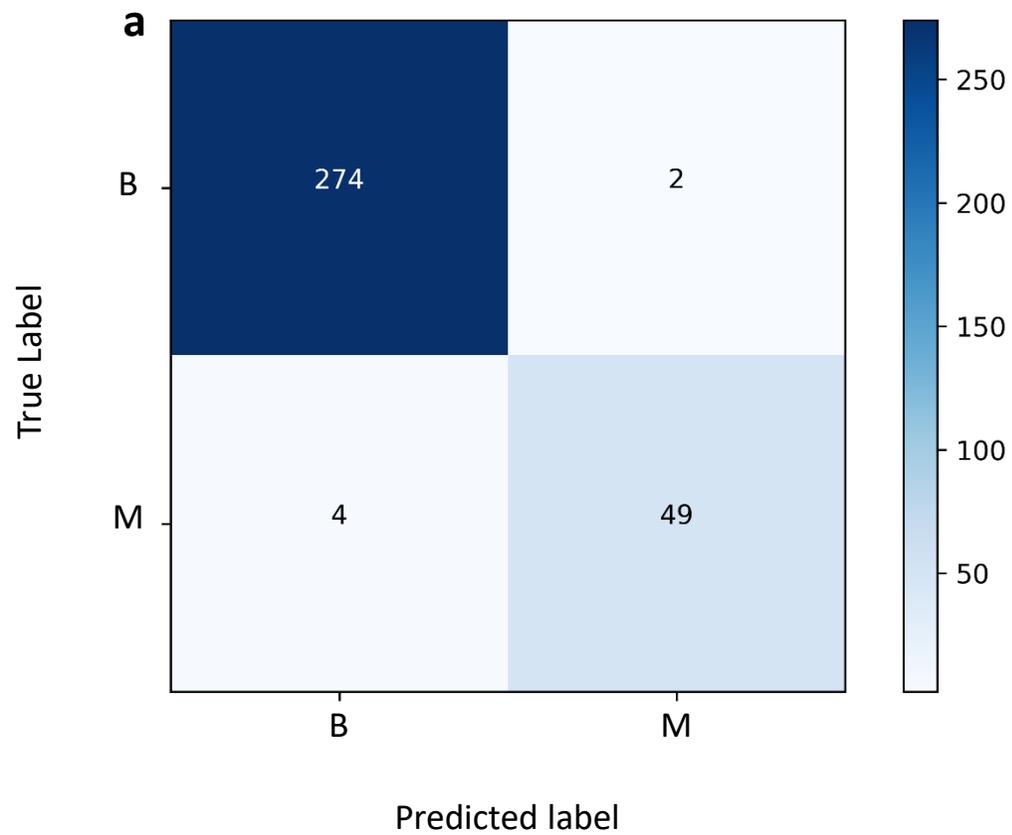 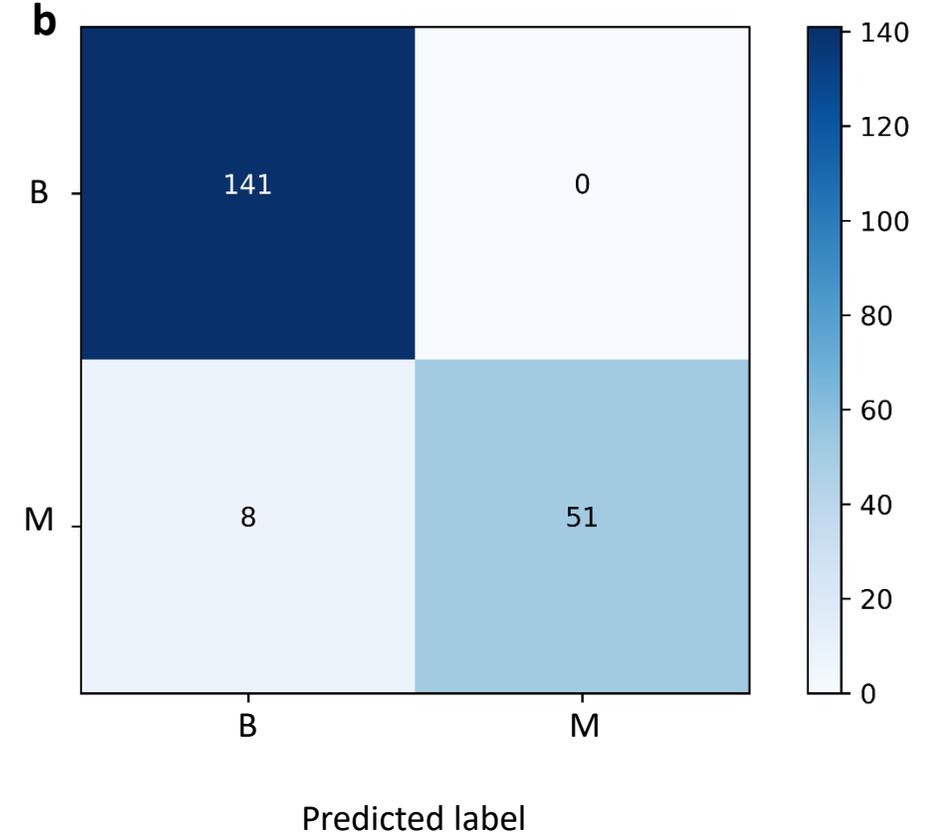

(**a**): Confusion matrix of HIENet$_{BC}$ for the two classes in the ten-fold cross-validation process, corresponding to Fig. 3e; (**b**): Confusion matrix of HIENet$_{BC}$ for the two classes in the external validation, corresponding to Fig. 7b.